%% file: root.tex
\documentclass[DTMColor]{ROB-New}
\usepackage{amssymb,amsfonts,amsmath,amscd}
\usepackage{xcolor}
\usepackage{dsfont}
\usepackage{dblfloatfix}    % To enable figures at the bottom of page
\usepackage{makecell}

\input{notation_header}

\begin{document}

\authormark{Sven Lilge and Timothy D. Barfoot}

\articletype{RESEARCH ARTICLE}

\jnlPage{}{}
\jyear{}
\jdoi{}

\title{Incorporating Control Inputs in Continuous-Time Gaussian Process State Estimation for Robotics}

\author[1]{Sven Lilge\hyperlink{corr}{*}}
\address[1]{University of Toronto Robotics Institute, University of Toronto, Toronto, Ontario, Canada.}

\author[1]{Timothy D. Barfoot}
\address{\hypertarget{corr}{*}Corresponding author. \email{sven.lilge@utoronto.ca}}

%\received{xx xxx xxx}
%\revised{xx xxx xxx}
%\accepted{xx xxx xxx}


\abstract{Continuous-time batch state estimation using Gaussian processes is an efficient approach to estimate the trajectories of robots over time.
	In the past, relatively simple physics-motivated priors have been considered for such approaches, using assumptions such as constant velocity or acceleration.
	This paper presents an approach to incorporating exogenous control inputs, such as velocity or acceleration commands, into the continuous Gaussian process state-estimation framework.
	It is shown that this approach generalizes across different domains in robotics, making it applicable to both the estimation of continuous-time trajectories for mobile robots and the estimation of quasi-static continuum robot shapes.
	Results show that incorporating control inputs leads to more informed priors, potentially requiring less measurements and estimation nodes to obtain accurate estimates.
	This makes the approach particularly useful in situations in which limited sensing is available. For example, in a mobile robot localization experiment with sparse landmark distance measurements and frequent odometry control inputs, our approach provides accurate trajectory estimates with root-mean-square errors around 3-4 cm and 4-5 degrees, even with time intervals up to five seconds between discrete estimation nodes, which significantly reduces computation time.}

\maketitle

%%%%%%%%%%%%%%%%%%%%%%%%%%%%%%%%%%%%%%%%%%%%%%%%%%%%%%%%%%%%%%%%%%%%%%%%%%%%%%%%
\section{Introduction}

To date, the majority of probabilistic state estimation approaches in robotics operate in discrete time.
However, such algorithms are usually pushed to their limits when considering high-rate sensor data, such as inertial measurement units (IMUs), or sensors that continuously collect data during motion, such as rolling-shutter cameras or spinning lidar and radar sensors. 
Continuous-time estimation techniques offer a solution to this problem by representing the state as a smooth trajectory over time, allowing inference of the state at any desired query time.
Unlike discrete-time methods, which need to include an additional state at every measurement time, continuous-time approaches can incorporate measurements at any time along the trajectory \cite{Furgale2012}.

\begin{figure}[t]
	\centering
	\includegraphics[width=1\linewidth]{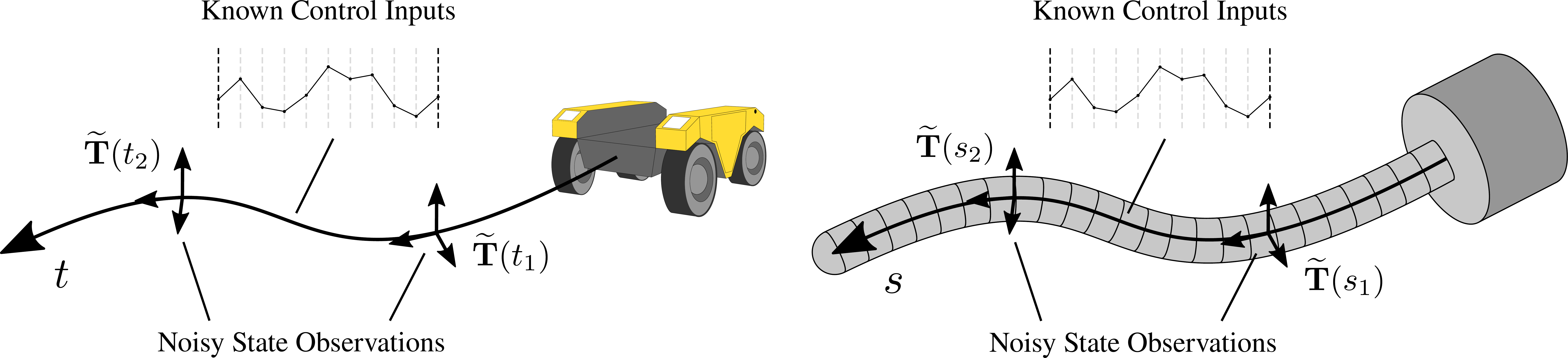}
	\caption{Our proposed method incorporates known control inputs into a continuous Gaussian process prior formulation, which is fused with noisy state observations. It is applicable to the estimation of both mobile robot continuous-time trajectories and continuum-robot shapes}
	\label{fig:figure1}
\end{figure}

Throughout this manuscript, we are specifically focusing on continuous-time batch state estimation using Gaussian processes (GPs) to represent the robot's state and uncertainty continuously over time \cite{Tong2013}.
This approach involves formulating a suitable prior distribution of continuous robot trajectories.
Current formulations are usually based on simple physics models, making specific assumptions about the motion of the robot.
In \cite{Anderson2015} a white-noise-on-acceleration (WNOA) prior is employed, assuming robot motion with constant velocity.
This approach has been extended to a white-noise-on-jerk (WNOJ) prior in \cite{Tang2019}, which assumes constant acceleration.
These formulations have been shown to also be applicable to the continuous estimation of the shape of continuum robots \cite{Lilge2022,Ferguson2024,Lilge2024}, substituting \textit{time} with \textit{arclength} and \textit{velocity} with \textit{strain}.
Fig.~\ref{fig:figure1} highlights this analogy.

For mobile robots, we seek to estimate their pose and velocity along a continuous trajectory, where the continuous variable is time.
For continuum robots, an analogous estimation problem can be formulated, in which we seek to estimate the robot's pose and strain along their continuous shape.
The continuous variable is now arclength and instead of estimating a trajectory over time, we are estimating a quasi-static shape in space.
In both cases, the robot's pose and its derivative are estimated along a continuous, smooth curve -- representing either a trajectory or shape -- by fusing the Gaussian process prior with noisy state observations of various types. For instance, in mobile robotics, noisy landmark observations are commonly used to relate the robot's pose and orientation to landmarks in the environment. In continuum robotics, electromagnetic tracking sensors might provide noisy direct measurements of the pose.

In many cases, the GP prior formulations represent the mobile robot motions or continuum robot shapes well.
However, besides the assumptions of constant velocity (or strain) or acceleration (equivalent to the strain derivative for continuum robots), no additional knowledge about the robot state is incorporated into the prior, such as exogenous control inputs to the system.
For mobile robots, these inputs could be known velocity or acceleration commands, while for continuum robots they could consist of known actuation forces and moments acting on their structure.
Incorporating such information directly into the GP formulation can lead to more informed prior distributions, representing the mobile robot trajectory or continuum robot shape more accurately.
Ultimately, this can mean that less measurements are required to accurately estimate the continuous robot state, when compared to the traditional prior formulations.

In this work, we are extending the existing WNOA Gaussian process prior formulation from \cite{Anderson2015} (for mobile robots) and \cite{Lilge2022} (for continuum robots) to include known exogenous control inputs.
We show how this approach can be applied to both mobile robot continuous-time trajectory estimation and continuum-robot shape estimation (see Fig.~\ref{fig:figure1}).

\subsection{Related Work}
In the following, we review commonly used continuous-time state estimation methods. For a more comprehensive introduction, we refer interested readers to a recent survey on these approaches \cite{Talbot2024}.

Initial approaches to infer the continuous state between two discrete states of a batch estimation approach relied on simple linear interpolation.
For instance, the works in \cite{Bosse2009,Bosse2012} assume constant velocities over short time periods to linearly interpolate the robot's pose for LiDAR Simultaneous Localization and Mapping (SLAM).
Similarly, \cite{Zhang2014,Hong2010} use linear pose interpolations to correct for motion distortion during LiDAR odometry.
While linear interpolation schemes are easy to implement and fast to compute, their accuracy is generally limited, as they do not necessarily correspond to the actual robot motion.

% Spline based methods
As an alternative, first parametric continuous-time state estimation approaches were proposed, using temporal basis functions to represent continuous and smooth robot trajectories \cite{Furgale2012}.
While initial investigations were limited to deal with states expressed in vector spaces, this approach has since been extended to Lie group formulations \cite{Lovegrove2013,Patron2015}.
Recent advancements mostly focus on increasing computational efficiency of these approaches, e.g., by deriving analytical Jacobians for Lie group formulations \cite{Sommer2020,Tirado2022,Li2023}.

% GP approach
A nonparametric approach, relying only on time as its sole parameter, was proposed in \cite{Tong2013, Tong2014}, using GPs to represent the continuous state in 2D and 3D.
Both of these works allow for a variety of different GP priors featuring relatively large and dense kernel matrices, ultimately limiting computational efficiency.
A more efficient GP prior was proposed in 
\cite{Barfoot2014}, which is derived from linear, time-varying stochastic differential equations driven by white noise, leading to an exactly sparse inverse kernel matrix by exploiting the Markov property. 
The proposed prior employs white noise on acceleration for 2D robot trajectories, leading to a constant velocity assumption.
The WNOA prior was extended to $SE(3)$ in \cite{Anderson2015}, while \cite{Tang2019} proposed an extension to consider white-noise-on-jerk prior formulations, assuming a constant acceleration instead of constant velocities.
In \cite{Wong2020a} the Singer prior is proposed, which introduces an additional parameter to the GP prior, representing latent accelerations.
Learning this parameter from data can potentially lead to priors that represent the robot's motion more accurately than the WNOA and WNOJ priors.
Lastly, derivations for formulations of the WNOA GP prior for general matrix Lie groups, beyond just $SE(3)$, are proposed in \cite{Dong2017,Dong2018}.
All of the aforementioned GP priors are relatively simple physically motivated formulations that do not consider exogenous control inputs.
Instead, velocity or acceleration inputs are often incorporated as measurements of the state \cite{Burnett2024, Burnett2024a}.

The continuous-time motion prior for the estimation of mobile robot trajectories was further repurposed to the continuous estimation of continuum robot shapes.
In \cite{Lilge2022} the WNOA prior was adopted to estimate the quasi-static continuous shape and strain of continuum robots, considering arclength as the continuous parameter instead of time.
This approach was extended to coupled robot topologies in \cite{Lilge2024}, while \cite{Ferguson2024} adopts a WNOJ prior to additionally estimate continuously applied forces to the continuum body.
A recent study compares the GP-based continuous-time estimation approach with parametric spline-based methods, noting that results are similar regarding the achieved accuracies and computation times \cite{Johnson2024}.

% GP for preintegration
Another approach to continuous state estimation using GPs is presented in \cite{Gentil2020} and \cite{Gentil2023}.
Here, GPs are used to pre-integrate high-rate IMU measurements, creating pseudo-measurements between two discretely estimated states.
The idea of pre-integrating high-rate measurements was first discussed in \cite{Lupton2011}, which uses numerical integration schemes, and is the backbone of many discrete-time batch estimation schemes \cite{Forster2015,Forster2016}.
Contrary to these traditional methods, the approach in \cite{Gentil2020} and \cite{Gentil2023} uses a continuous GP latent space representation fitted to measurements, which is then integrated in closed form (i.e., Bayesian quadrature).
This results in a smooth pseudo-measurement, which is then used as a factor between two discrete states in a batch estimation approach.
Unlike the continuous-time estimation methods discussed before, this work does not feature an underlying continuous motion prior and carries out estimation purely in discrete time after pre-integration.
It is thus unable to estimate the continuous mean and covariance of a robot's trajectory and could fail if the high-rate measurements between two discrete states drop out.
A similar pre-integration approach is presented in \cite{Cioffi2023}.
Here, conventional on-manifold pre-integration \cite{Forster2016} is combined with a data-driven dynamics model, to incorporate dynamic motion model factors into the discrete-time batch estimation of drone trajectories.

To conclude, the main limitations of current approaches lie in the simplicity of the physical assumptions underlying most continuous-time state estimation methods, such as constant velocity or constant acceleration.
While these assumptions work well if the robot motion is close to these conditions, they fail to adequately represent more complex or varying motion.
Pre-integration methods partially offer a solution by capturing robot motion through pre-integrated factors considering high-rate measurements.
However, they revert back to a discrete-time estimation problem at runtime, sacrificing the ability to continuously query the state at any desired time.

Consequently, a key question remains open: how can more intricate physics models, especially ones incorporating exogenous control inputs, be integrated into continuous-time state estimation frameworks?
Addressing this challenge is crucial for formulating priors that more accurately represent the actual motion of robots.
This issue is also identified as an open research question in a recent survey on continuous-time state estimation methods \cite{Talbot2024}.
Tackling this problem has the potential to enhance both the accuracy and applicability of continuous-time state estimation methods.

\subsection{Contributions}

This work presents a direct extension of the established WNOA Gaussian process prior \cite{Anderson2015,Lilge2022} to include exogenous control inputs.
The derivations presented throughout this manuscript allow for the inclusion of inputs on both the velocity and acceleration.
To the best of the authors' knowledge, this work presents the first GP-based continuous-time estimation approach that combines a physically motivated motion prior with such control inputs.
By combining ideas from the pre-integration literature \cite{Gentil2020,Gentil2023} with previously proposed continuous-time motion priors \cite{Anderson2015,Lilge2022}, a single factor between two discrete robot states is derived, which can then be used in a batch estimation approach.

While our method shares similarities with existing approaches in the current literature, there are some striking differences that set it apart.
As discussed in the previous section, continuous-time state estimation methods for robotics are traditionally based on temporal basis functions, such as splines, or GPs \cite{Talbot2024,Johnson2024}.
While spline formulations are not directly tied to motion process models, GPs are motivated by the underlying physics of robot motion, theoretically allowing the inclusion of complex process models and control inputs.
However, previous GP-based methods have been limited to simple motion models such as constant velocity or constant acceleration \cite{Anderson2015,Tang2019}.

Our approach makes contributions to the field by introducing a GP prior capable of representing more complex robot motions defined by exogenous control inputs.
This is the first time such an extension has been proposed.
Consequently, our priors are more informed, more general, and potentially more accurate. Similar to pre-integration methods \cite{Gentil2020,Gentil2023}, which pre-integrate IMU measurements to compute relative pseudo-measurement factors, our motion prior factors describe the relative motion between discrete states in time.
However, unlike pre-integration methods, which estimate the states at discrete times and are unable to query it at arbitrary times after computation of the factors, our GP-based approach allows straightforward interpolation of the mean and covariance of the robot’s state at any query time.
This enables the recovery of the local shape and uncertainty of the trajectory between discrete states.
Additionally, in the case in which no known inputs are available, the proposed method gracefully falls back to the underlying WNOA motion prior for estimation and interpolation.

In conclusion, this work advances continuous-time state estimation, introducing a GP-based framework that incorporates process models considering exogenous control inputs, potentially paving the way for more versatile continuous-time estimation approaches in robotics.
Throughout this manuscript, we derive formulations that consider continuous control inputs in $SE(3)$, allowing us to approximate the resulting GP prior in closed form.
The proposed approach is applied to both the estimation of continuous mobile robot trajectories and quasistatic continuum robot shapes, highlighting advantages over the existing GP motion priors.
We show that considering control inputs is particularly useful when the available measurement data is sparse.

\section{Continuous State Estimation Using Gaussian Process Regression with Inputs}

In the following, we outline derivations for our new GP prior formulation considering control inputs.
We further briefly discuss how the resulting prior can be used in a \textit{maximum a posteriori} (MAP) batch estimation scheme, but refer the reader to \cite{Anderson2015,Lilge2022} for additional details.

We note that the following derivations are written for mobile robot trajectory estimation, featuring time as the continuous independent variable as well as pose and generalized velocity (i.e., twist) as the robot's state variables.
However, analogous formulations can be derived for quasi-static continuum robot shape estimation problems, replacing time with arclength and velocity with strain (see \cite{Lilge2022} for details of this analogy).

\subsection{Gaussian Process Prior}

Throughout the following, we are using the notation established in \cite{Barfoot2024}.
We describe the continuous-time mobile robot state with the following time-varying stochastic differential equations:
\begin{subequations}
	\begin{align} \label{eq:motion_ode}
		\dot{\mbf{T}}(t) &= (\underbrace{\mbs{\varpi}_\mathrm{bias}(t) + \mbf{v}_\mathrm{in}(t)}_{\mbs{\varpi}(t)})^\wdg\mbf{T}(t), \\
		\dot{\mbs{\varpi}}_\mathrm{bias}(t) &= \mbf{a}_\mathrm{in}(t) + \mbf{w}(t), \\ \mbf{w}(t) &\sim \mathcal{GP}(\mbf{0},\mbf{Q}_c(t-t')).
	\end{align}
\end{subequations}
These equations describe how the robot's pose $\mbf{T}(t) \in SE(3)$ and body-centric velocity $\mbs{\varpi}(t) \in \mathds{R}^6$ evolve over time.
In contrast to the traditional WNOA equations from \cite{Anderson2015}, we consider two additional inputs terms, one for velocity $\mbf{v}_\mathrm{in}(t) \in \mathds{R}^6$  and one for acceleration inputs $\mbf{a}_\mathrm{in}(t) \in \mathds{R}^6$.
The overall robot body-centric velocity $\mbs{\varpi}(t)$ now consists of both the velocity inputs $\mbf{v}_\mathrm{in}(t) \in \mathds{R}^6$ and a velocity bias term $\mbs{\varpi}_\mathrm{bias}(t)$.
This bias term describes motion that is not a direct result of the velocity inputs.
Its derivative is driven by a white-noise GP $\mbf{w}(t)$ with positive-definite power-spectral density matrix $\mbf{Q}_c$.
In addition, we consider potential acceleration inputs $\mbf{a}_\mathrm{in}(t)$ as part of this derivative.

Given the stated equations, both inputs $\mbf{v}_\mathrm{in}(t)$ and $\mbf{a}_\mathrm{in}(t)$ are independent from each other (i.e., $\mbf{a}_\mathrm{in}(t)$ is not the derivative of $\mbf{v}_\mathrm{in}(t)$).
In theory, both inputs can be applied simultaneously, resulting in robot motion described by a superposition of velocity and acceleration inputs.
In practice, we expect to use either velocity or acceleration inputs, with either $\mbf{v}_\mathrm{in}(t)$ or $\mbf{a}_\mathrm{in}(t)$ set to zero.
When both inputs are zero, the equations simplify to the traditional WNOA case discussed in \cite{Anderson2015}.

\begin{figure}[t]
	\centering
	\includegraphics[width=0.7\linewidth]{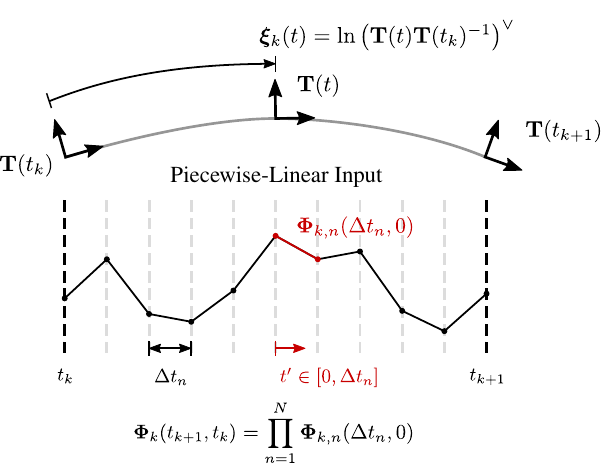}
	\caption{Top: Definition of local pose variables, $\mbs{\xi}_k(t)$, between two discrete robot states. Bottom: Example of piecewise-linear inputs between two discrete robot states. The overall transition function between the two states $\mbf{\Phi}_k(t_{k+1},t_k)$ is a product of the individual transition functions for each piecewise-linear segment}
	\label{fig:discrete_states}
\end{figure}

Following our previous work, we use a series of local Gaussian processes to approximate the nonlinear stochastic differential equations above.
We discretize the continuous robot state into $K$ discrete states at times $t_k$.
Between each pair of neighbouring estimation times, $t_k$ and $t_{k+1}$, local pose variables are defined in the Lie algebra $\mbs{\xi}_k(t) \in \mathfrak{se}(3)$ (see Fig.~\ref{fig:discrete_states}, top).
We can now write
\begin{align}
	\mbs{\xi}_k(t) &= \ln\left(\mbf{T}(t)\mbf{T}(t_k)^{-1}\right)^\vee \label{eq:local_1},\\
	\dot{\mbs{\xi}}_k(t) &= \underbrace{\mbs{\mathcal{J}}_k^{-1}\mbs{\varpi}_\mathrm{bias}(t)}_{\mbs{\psi}_k(t)} + {\mbs{\mathcal{J}}}_k^{-1}\mbf{v}_{k,\mathrm{in}}(t),\label{eq:local_2} \\
	\dot{\mbs{\psi}}_k(t) &= \dot{\mbs{\mathcal{J}}}_k^{-1}\mbs{\mathcal{J}}_k\mbs{\psi}_k(t) + \mbs{\mathcal{J}}_k^{-1}\left(\mbf{a}_{k,\mathrm{in}}(t) + \mbf{w}_k(t)\right),
\end{align}
where $\mbs{\mathcal{J}}_k~=~\mbs{\mathcal{J}}(\ln\left(\mbf{T}(t)\mbf{T}(t_k)^{-1}\right)^\vee)$ is the left Jacobian of $SE(3)$ \cite{Barfoot2024} and inputs $\mbf{v}_{k,\mathrm{in}}(t)$, $\mbf{a}_{k,\mathrm{in}}(t)$, as well as noise $\mbf{w}_k(t)$ are now defined on the local time interval.
Considering the local Markovian state $\mbs{\gamma}_k(t) = \bbm\mbs{\xi}_k(t)^T & \mbs{\psi}_k(t)^T\ebm^T$, we can write the following time-varying stochastic differential equations:
\begin{align}
	\bbm \dot{\mbs{\xi}}_k(t) \\ {\dot{\mbs{\psi}}_k(t)}  \ebm = \bbm \mbf{0} & \mbf{1} \\ \mbf{0} & \dot{\mbs{\mathcal{J}}}_k^{-1}\mbs{\mathcal{J}}_k \ebm  \bbm \mbs{\xi}_k(t) \\ {\mbs{\psi}_k(t)} \ebm + \mbs{\mathcal{J}}_k^{-1}\bbm \mbf{v}_{k,\mathrm{in}}(t) \\ \mbf{a}_{k,\mathrm{in}}(t) \ebm + \bbm \mbf{0} \\ \mbs{\mathcal{J}}_k^{-1} \ebm \mbf{w}_k(t),
\end{align}
We can further simplify this expression as
\begin{align}
	\bbm \dot{\mbs{\xi}}_k(t) \\ {\dot{\mbs{\psi}}_k(t)}  \ebm \approx \underbrace{\bbm \frac{1}{2}\mbf{v}_{k,\mathrm{in}}(t)^\curlywedge & \mbf{1} \\ \frac{1}{2}\mbf{a}_{k,\mathrm{in}}(t)^\curlywedge & -\frac{1}{2}\mbf{v}_{k,\mathrm{in}}(t)^\curlywedge \ebm}_{\mbf{A}_k(t)} \bbm \mbs{\xi}_k(t) \\ {\mbs{\psi}_k(t)} \ebm + \bbm \mbf{v}_{k,\mathrm{in}}(t) \\ \mbf{a}_{k,\mathrm{in}}(t) \ebm + \bbm \mbf{0} \\ \mbf{1} \ebm \mbf{w}_k(t),
\end{align}
while using the approximations $\mbs{\mathcal{J}}_k^{-1} \approx \mbs{1} - \frac{1}{2}\mbs{\xi}_k(t)^\Wdg$ and $\mbs{\mathcal{J}}_k \approx \mbs{1} + \frac{1}{2}\mbs{\xi}_k(t)^\Wdg$, and assuming that $\mbs{\xi}_k(t)$ and $\mbs{\varpi}_\mathrm{bias}(t)$ are small.
The above expression can now be integrated stochastically in closed form to obtain a GP prior (i.e., kernel function) with mean $\mbs{\check{\gamma}}_k(t)$ and covariance $\check{\mbf{P}}(t)$:
\begin{align}
	\mbs{\check{\gamma}}_k(t) &= \mbs{\Phi}_k(t,t_k) \pri{\mbs{\gamma}}_k(t_k) + \int_{t_k}^{t}\mbs{\Phi}_k(t,s)\bbm \mbf{v}_{k,\mathrm{in}}(s) \\ \mbf{a}_{k,\mathrm{in}}(s) \ebm ds, \label{eq:prior_mean} \\
	\check{\mbf{P}}(t) &= \mbs{\Phi}_k(t,t_k) \pri{\mbf{P}}(t_k)  \mbs{\Phi}_k(t,t_k)^T + \mbf{Q}(t - t_k). \label{eq:prior_cov}
\end{align}
Here, ${\mbs{\Phi}}_k(t,t_0)$ is the transition function between two times. It is computed by solving the initial value problem of the differential equation
\begin{align}
	\dot{\mbs{\Phi}}_k(t,t_0) &= \mathbf{A}_k(t) {\mbs{\Phi}}_k(t,t_0),\qquad 
	\mbs{\Phi}_k(t_0,t_0) = \mathbf{1} \label{eq:ode_phi}.
\end{align}
Additionally, $\mathbf{Q}(t - t_k)$ is the covariance accumulated between two times, which can be computed as
\begin{align}
	\mathbf{Q}(t - t_k) = \int_{t_k}^t{\mbs{\Phi}}_k(t,s)\bbm \mathbf{0} \\ \mathbf{1} \ebm \mathbf{Q}_c \bbm \mathbf{0} & \mathbf{1} \ebm{\mbs{\Phi}}_k(t,s)^Tds. \label{eq:acc_cov}
\end{align}

\subsection{Prior Error Term}
Next, we derive an error term, evaluating how well a given state aligns with the derived GP prior.
The general motion-prior error term for GP regression is
\begin{align}
	\mbf{e}_{p,k} = &\left( \mbs{\gamma}_k(t_{k+1}) - \pri{\mbs{\gamma}}_k(t_{k+1})\right) - \mbs{\Phi}_k(t_{k+1},t_{k}) \left( \mbs{\gamma}_k(t_{k}) - \pri{\mbs{\gamma}}_k(t_{k})\right).
\end{align}
Substituting our expression for the state mean in \eqref{eq:prior_mean} leads to
\begin{align}
	\mbf{e}_{p,k} =  &\mbs{\gamma}_k(t_{k+1}) - \mbs{\Phi}_k(t_{k+1},t_{k}) \mbs{\gamma}_k(t_{k}) - \int_{t_{k}}^{t_{k+1}}\mbs{\Phi}_k(t_{k+1},s)\bbm \mbf{v}_{k,\mathrm{in}}(s) \\ \mbf{a}_{k,\mathrm{in}}(s) \ebm ds.
\end{align}
Using \eqref{eq:local_1} and \eqref{eq:local_2}, we can further express this error term using our global variables as
\begin{align}
	\mbf{e}_{p,k} &=  \bbm \ln\left(\mbf{T}(t_{k+1})\mbf{T}(t_{k})^{-1}\right)^\vee \\  \mbs{\mathcal{J}}\left(\ln\left(\mbf{T}(t_{k+1})\mbf{T}(t_{k})^{-1}\right)^\vee\right)^{-1}\mbs{\varpi}_\mathrm{bias}(t_{k+1}) \ebm - \mbs{\Phi}_k(t_{k+1},t_{k}) \bbm \mbf{0} \\  \mbs{\varpi}_\mathrm{bias}(t_{k}) \ebm \label{eq:prior_error}\\
	& \qquad - \int_{t_{k}}^{t_{k+1}}\mbs{\Phi}_k(t_{k+1},s)\bbm \mbf{v}_{k,\mathrm{in}}(s) \\ \mbf{a}_{k,\mathrm{in}}(s) \ebm ds\notag
\end{align}
Using this prior error, we can construct the following squared-error cost term to represent its negative log-likelihood:
\begin{align}
	J_{p,k} = \frac{1}{2}\mbf{e}^T_{p,k}\mathbf{Q}^{-1}_{k}\mbf{e}_{p,k},
\end{align}
with $\mbf{Q}_{k} = \mbf{Q}(t_{k+1}-t_{k})$.
Each of the prior cost terms is a binary factor between two discrete states if thought of as a factor-graph representation.
It expresses how close the two corresponding states are to the assumption of the motion prior, considering the incorporated control inputs.

\subsection{Maximum A Posteriori Batch State Estimation}

We can use these cost terms in a MAP batch state estimation approach, where they are weighed off against any noisy state observations and measurements.
This allows us to obtain a GP posterior $\hat{\mbs{\gamma}}(t_k)$ of the state at discrete times $t_k$.
We can again use \eqref{eq:local_1} and \eqref{eq:local_2} to obtain a posterior estimate of our global state variables $\hat{\mbf{T}}(t_k)$ and $\hat{\mbs{\varpi}}(t_k)$ at these times. 
We refer to \cite{Anderson2015} and \cite{Lilge2022} for details on the implementation of the MAP batch state estimation approach.

It is noted that during the MAP optimization, the transition function in \eqref{eq:ode_phi}, the accumulated covariance in \eqref{eq:acc_cov}, and the integral term in \eqref{eq:prior_error} only have to be computed during the first iteration and can be held constant afterwards, as they do not depend on the state variables themselves.
This significantly reduces the computational burden of the newly introduced derivations.

\subsection{Gaussian Process Interpolation}

Lastly, we are interested in interpolating the GP posterior at any arbitrary time $\tau$ along the continuous trajectory.
For this, we start with the general interpolation equations for the GP mean stated in \cite[Sec.~11.3]{Barfoot2024}:
\begin{align}
	\hat{\mbs{\gamma}}_k(\tau) =  \check{\mbs{\gamma}}_k(\tau) &+  \mbs{\Lambda}_k(\tau)\left(\hat{\mbs{\gamma}}_k(t_k) - \check{\mbs{\gamma}}_k(t_k)\right) + \mbs{\Psi}_k(\tau)\left( \hat{\mbs{\gamma}}_k(t_{k+1})  -  \check{\mbs{\gamma}}_k(t_{k+1})\right),
\end{align}
with
\begin{align}
	\mbs{\Lambda}_k(\tau) &= \mbs{\Phi}_k(\tau,t_k) - \mbf{Q}_\tau\mbs{\Phi}_k(t_{k+1},\tau)^T\mbf{Q}^{-1}_{k}\mbs{\Phi}_k(t_{k+1},t_k), \\
	\mbs{\Psi}_k(\tau) &= \mbf{Q}_\tau\mbs{\Phi}_k(t_{k},\tau)^T\mbf{Q}^{-1}_{k},
\end{align}
where $\mbf{Q}_\tau = \mbf{Q}(\tau-t_{k})$.
In our case, we can rewrite these equations using the expression for our prior mean in \eqref{eq:prior_mean} as
\begin{align}
	\hat{\mbs{\gamma}}_k(\tau) &= \int_{t_k}^{\tau}\mbs{\Phi}_k(\tau,s)\bbm \mbf{v}_{k,\mathrm{in}}(s) \\ \mbf{a}_{k,\mathrm{in}}(s) \ebm ds + \mbs{\Lambda}_k(\tau)\hat{\mbs{\gamma}}_k(t_k) + \mbs{\Psi}_k(\tau)\hat{\mbs{\gamma}}_k(t_{k+1}) \label{eq:interpolation} \\  & \qquad - \mbs{\Psi}_k(\tau)\int_{t_{k}}^{t_{k+1}}\mbs{\Phi}(t_{k+1},s)\bbm \mbf{v}_{k,\mathrm{in}}(s) \\ \mbf{a}_{k,\mathrm{in}}(s)\ebm ds. \notag
\end{align}
Now, we can solve for $\hat{\mathbf{T}}(\tau)$ and $\hat{\mbs{\varpi}}(\tau)$ considering the mapping between global and local pose variables in \eqref{eq:local_1} and \eqref{eq:local_2}.
The covariance of the posterior GP can be interpolated similarly.
Derivations are omitted for the sake of space, but details can be found in \cite[Sec.~11.3]{Barfoot2024} and \cite{Anderson2017a}.
This interpolation scheme can further be used during the MAP optimization to incorporate measurement terms at query times different from our discrete times $t_k$ \cite{Burnett2024a}.

Although querying the trajectory remains $O(1)$, it may be computationally more expensive than in the WNOA case \cite{Anderson2015}, depending on the inputs $\mbf{v}_{k,\mathrm{in}}(t)$ and $\mbf{a}_{k,\mathrm{in}}(t)$.
This increase arises from two factors: (1) the transition function ${\mbs{\Phi}}_k(t,t_0)$ and accumulated covariance $\mbf{Q}(t-t_0)$ may require more computation, and (2) the integral terms in \eqref{eq:interpolation} add additional overhead.

This completes the general derivations of our GP prior formulation considering inputs.
We note that the newly introduced expressions are not analytically solvable in the general case.
In the following, we will discuss the particular example of piecewise-linear inputs, for which analytical approximations of all necessary formulations can be derived.

\subsection{Example of Piecewise-Linear Inputs}

We discretize the inputs between two discrete states at times $t_k$ and $t_{k+1}$ into $n\in \left\{1, ..., N\right\}$ segments, each represented by a piecewise-linear function (see Fig.~\ref{fig:discrete_states}, bottom).
The inputs over each individual segment are now defined on a local interval $t' \in [0,\Delta t_n]$ as
\begin{align}
	\mbf{v}_{k,n,\mathrm{in}}(t') &= \mbf{v}_{k,n,\mathrm{in}}(0) + \frac{\mbf{v}_{k,n,\mathrm{in}}(\Delta t_n) - \mbf{v}_{k,\mathrm{in}}(0)}{\Delta t_n}t',	\label{eq:v_lin} \\
	\mbf{a}_{k,n,\mathrm{in}}(t') &= \mbf{a}_{k,n,\mathrm{in}}(0) + \frac{\mbf{a}_{k,n,\mathrm{in}}(\Delta t_n) - \mbf{a}_{k,n,\mathrm{in}}(0)}{\Delta t_n}t', \label{eq:a_lin}
\end{align}
where $\Delta t_n = ({t_{k+1} - t_k})/{N}$.
Given these expressions for our inputs, we can now find the piecewise transition function between two times $t'_0$ and $t'$ for each individual segment.
Given linear inputs in the form of \eqref{eq:v_lin} and \eqref{eq:a_lin}, the system matrix in \eqref{eq:ode_phi} can be written as a linear function itself on the local interval such that
\begin{align}
	\dot{\mbs{\Phi}}_{k,n}(t',t'_0) &= \mathbf{A}_{k,n}(t') {\mbs{\Phi}}_{k,n}(t',t'_0),\qquad 
	\mbs{\Phi}_{k,n}(t'_0,t'_0) = \mathbf{1}, \label{eq:ode_phi_local}
\end{align}
with $\mbf{A}_{k,n}(t') = \mbf{B}_{k,n} + \mbf{C}_{k,n}t'$.
Solving this ODE allows us to describe the transition function on the local interval $t' \in [0,\Delta t_n]$.
While there exists no closed-form solution to this differential equation, we can use the Magnus expansion \cite{Blanes2009} to approximate the piecewise transition function for each segment $n$.
The Magnus expansion expresses the result of a linear ODE, such as the one in \eqref{eq:ode_phi_local}, as an exponential series:
\begin{align}
	\mbs{\Phi}_{k,n}(t',t'_0) = \exp\left(\mbs{\Omega}(t)\right), \qquad \mbs{\Omega}(t') = \sum_{i=1}^{\infty}\mbs{\Omega}_i(t').
\end{align}
The first three terms of $\mbs{\Omega}(t)$ are given as
\begin{align}
	\mbs{\Omega}_1(t') &= \int_{t'_0}^{t'}\mathbf{A}_{k,n}(t'_1)dt'_1, \\
	\mbs{\Omega}_2(t') &= \frac{1}{2}\int_{t'_0}^{t'}\int_{t'_0}^{t'_1}\left[\mathbf{A}_{k,n}(t'_1),\mathbf{A}_{k,n}(t'_2)\right]dt'_2 dt'_1, \\
	\mbs{\Omega}_3(t') &= \frac{1}{6}\int_{t'_0}^{t'}\int_{t'_0}^{t'_1}\int_{t'_0}^{t'_2}\left(\left[\mathbf{A}_{k,n}(t'_1),\left[\mathbf{A}_{k,n}(t'_2),\mathbf{A}_{k,n}(t'_3)\right]\right] + \right. \notag \\ & \qquad \qquad \qquad \qquad \quad \left. \left[\mathbf{A}_{k,n}(t'_3),\left[\mathbf{A}_{k,n}(t'_2),\mathbf{A}_{k,n}(t'_1)\right]\right]\right)dt'_1dt'_2dt'_3,
\end{align}
where $\left[\mathbf{A},\mathbf{B}\right] = \mathbf{AB} - \mathbf{BA}$ is the usual Lie bracket.
Because $\mbf{A}_{k,n}(t')$ is linear, given our piecewise-linear inputs, we can compute these expressions in closed form.
Using these first three terms of the Magnus expansion, we now have
\begin{align}
	\mbs{\Phi}_{k,n}(t',t'_0) \approx &\exp \Bigl(\mbf{B}_{k,n}(t'-t'_0) + \frac{1}{2}\mbf{C}_{k,n}(t'^2-t_0'^2) + \frac{1}{12}\left[\mbf{C}_{k,n},\mbf{B}_{k,n}\right](t'-t'_0)^3  \label{eq:magnus}\\ & \qquad \qquad + \frac{1}{240}\left[\mbf{C}_{k,n}\left[\mbf{C}_{k,n},\mbf{B}_{k,n}\right]\right](t'-t'_0)^5\Bigr). \notag
\end{align}
Intuitively, including more terms in the Magnus expansion would improve the approximation of our transition function, but we find the first three terms sufficiently accurate.
The total resulting transition function between the two discrete times $t_k$ and $t_{k+1}$ can now be computed by concatenating the individual piecewise ones such that
\begin{align}
	\mbs{\Phi}_k(t_{k+1},t_k) = \prod_{n=1}^N\mbs{\Phi}_{k,n}(\Delta t_n,0).
\end{align}
In order to fully express the prior mean and covariance, we need to compute additional integral terms.
For the mean in \eqref{eq:prior_mean}, we compute
\begin{align}
	\int_{t_k}^{t}\mbs{\Phi}_k(t,s)\bbm \mbf{v}_{k,\mathrm{in}}(s) \\ \mbf{a}_{k,\mathrm{in}}(s) \ebm ds 
	&= \sum_{n=1}^{N}\int_{t_{n-1}}^{\mathrm{min}(t_n,t)}\mbf{\Phi}_k\left(t,s\right)\bbm \mbf{v}_{k,\mathrm{in}}(s) \\ \mbf{a}_{k,\mathrm{in}}(s) \ebm ds,\\
	&= \sum_{n=1}^{N}\int_{t_{n-1}}^{\mathrm{min}(t_n,t)}\underbrace{\mbf{\Phi}_k\left(t,t_{N-1}\right)\cdots}_\mathrm{constant}\mbf{\Phi}_k\left(t_n,s\right)\bbm \mbf{v}_{k,\mathrm{in}}(s) \\ \mbf{a}_{k,\mathrm{in}}(s) \ebm ds, \label{eq:integral} 
\end{align}
with
\begin{align}
	t_n &= t_k + \frac{n}{N}\left(t_{k+1} - t_k\right).
\end{align}
Here, we are computing the whole integral by splitting it up into several smaller intervals, defined by the discretization of the piecewise-linear input as shown in Fig.~\ref{fig:discrete_states}.
By writing out the transition function in these integrals as a product of the individual transition functions of each segment, we see that only one of those terms depends on the integration variable $s$, allowing us to pull the remaining terms out of the integral.
Given the analytical approximation from \eqref{eq:magnus} and the expression for the piecewise-linear inputs in \eqref{eq:v_lin} and \eqref{eq:a_lin}, we can now solve the integral in closed form.
We omit the details of this derivation for brevity.
We further note that we truncate the computed integral after the fifth-order terms, which we find sufficiently accurate for implementation.

The same strategy can be used to compute the integral for the accumulated covariance $\mbf{Q}(t - t_k)$ in \eqref{eq:acc_cov}, which we need for the prior covariance in \eqref{eq:prior_cov}:
\begin{align}
	\mbf{Q}(t - t_k) &= \int_{t_k}^{t}\mbs{\Phi}_k(t,s)\bbm \mbf{0} \\ \mbf{1} \ebm\mbf{Q}_c\bbm \mbf{0} & \mbf{1} \ebm\mbs{\Phi}_k(t,s)^T ds, \\
	&= \sum_{n=1}^{N}\int_{t_{n-1}}^{\mathrm{min}(t_n,t)}\mbs{\Phi}_k(t,s)\bbm \mbf{0} \\ \mbf{1} \ebm\mbf{Q}_c\bbm \mbf{0} & \mbf{1} \ebm\mbs{\Phi}_k(t,s)^T ds, \\
	&= \sum_{n=1}^{N}\int_{t_{n-1}}^{\mathrm{min}(t_n,t)}
	\underbrace{\mbf{\Phi}_k\left(t,t_{N-1}\right)\cdots}_\mathrm{constant}\mbf{\Phi}_k\left(t_n,s\right)\bbm \mbf{0} \\ \mbf{1} \ebm\mbf{Q}_c\bbm \mbf{0} & \mbf{1} \ebm\mbf{\Phi}_k\left(t_n,s\right)^T \underbrace{\cdots\mbf{\Phi}_k\left(t,t_{N-1}\right)^T}_\mathrm{constant} ds.
\end{align}
Using the same strategy as above, we see that several constant terms can again be pulled out of the integral and we can solve the remaining expression using the analytical approximation from \eqref{eq:magnus}.
We again omit the detailed steps for solving this integral and truncate the result after the fifth order during implementation, which we find sufficiently accurate.

Given the above derivations, further simplifications can be made if the inputs between two times are constant, in which case $\mbf{A}_{k,n}(t') = \mathbf{B}_{k,n}$, or zero, in which case the derivations will simplify to the constant-velocity prior discussed in \cite{Anderson2015}.
\begin{figure}[t]
	\centering
	\includegraphics[width=1\linewidth]{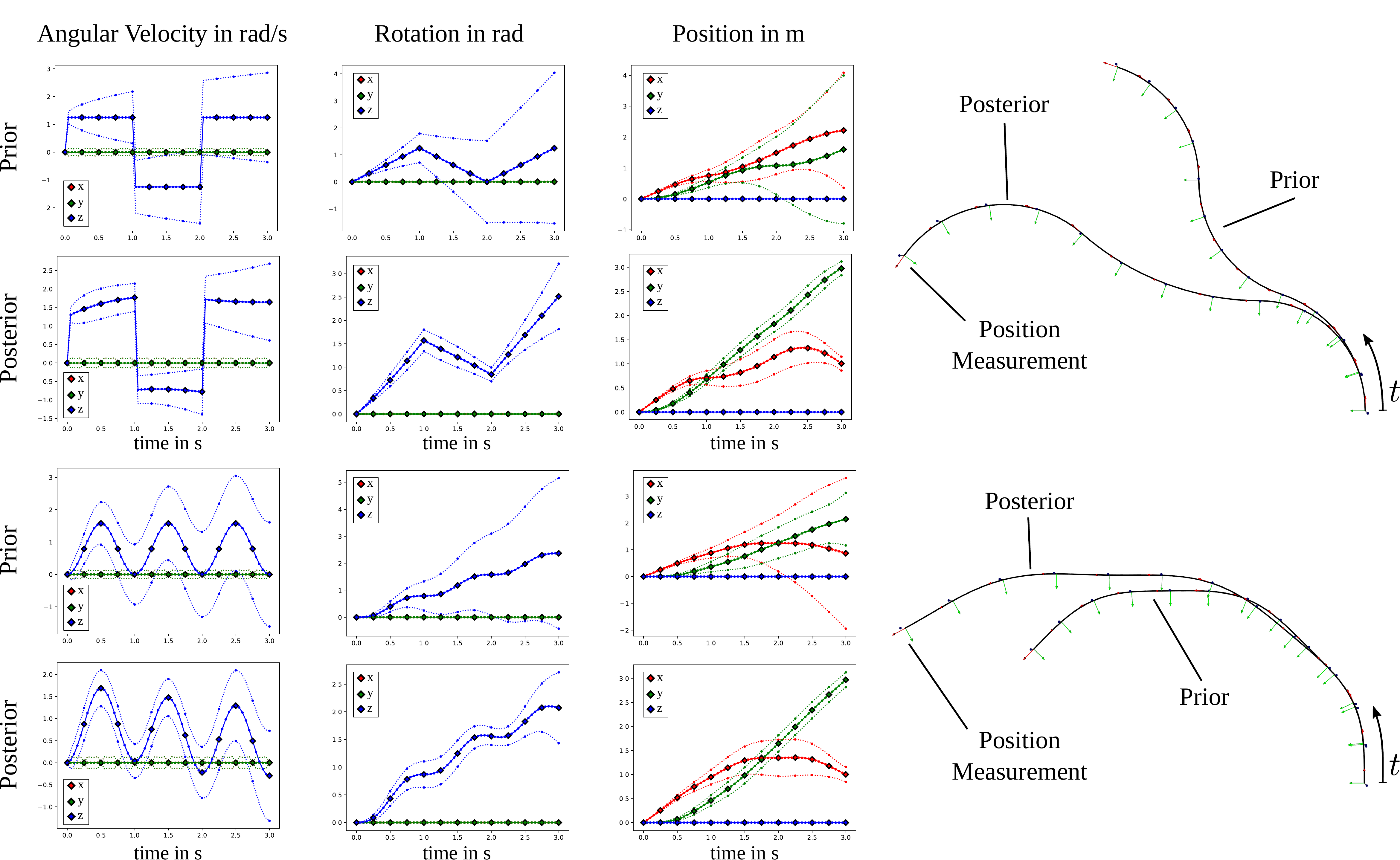}
	\caption{Example scenarios using the proposed GP prior formulation, featuring angular velocity inputs (top) and angular acceleration inputs (bottom). Both scenarios highlight the resulting prior as well as the posterior, when considering an additional position measurement. In each case, the angular velocities as well as the rotation and position are plotted over time, including mean and $3\sigma$-covariance envelopes and using red, green, and blue colours for the $x$, $y$, and $z$-components, respectively. Estimation nodes are highlighted with diamonds. Additional renderings of the resulting trajectories are depicted on the right}
	\label{fig:sim_examples}
\end{figure}

As mentioned previously, the transition function, accumulated covariances and integrals discussed above generally only have to be computed once.
Afterwards, they can be kept constant during the MAP optimization, as they do not depend on the state variables themselves.
Nevertheless, when querying the posterior GP after convergence at arbitrary times, some transition functions, accumulated covariances, and integrals must be re-evaluated, which is now more computationally expensive than in the WNOA case \cite{Anderson2015}.
However, by storing most of the intermediate results of the initial computations, some terms can be reused for this interpolation, which leads to an increased efficiency.
This is particularly useful if the interpolation times coincide with the piecewise-linear discretization shown in Fig.~\ref{fig:discrete_states}.

\subsection{Examples}

Fig.~\ref{fig:sim_examples} shows two example scenarios using the proposed GP prior formulation with inputs.
Both examples feature robot trajectories over three seconds, while considering a known initial pose and forward velocity of 1 m/s.

The prior of the first example uses piecewise-constant angular velocity inputs featuring discontinuities, leading to a prior shape composed of three concatenated constant-curvature arcs.
The second example considers a sinusoidal angular acceleration input, approximated via piecewise-linear inputs, leading to the depicted sinusoidal velocity profile.
For both examples, we show the employed prior as well as the resulting posterior, when an additional position measurement at the end of the trajectory is considered.
It can be seen that after fusing with the measurement, the posterior maintains the local shape of the prior in both cases.
Further, the robot's velocity and pose can be accurately recovered along the continuous trajectory using GP interpolation.
Thus, even though both examples consider the same position measurement, vastly different posterior trajectories are achieved.
We further note that the posterior is able to maintain the discontinuities in the velocity profile of the first example.
This would not be possible when using the conventional WNOA prior, which could consider the velocity inputs as measurements, as it forces smooth and continuous velocity profiles.

\section{Mobile Robot Trajectory Estimation}

In the following, we are evaluating our proposed method using a mobile robot trajectory estimation experiment.

\subsection{Dataset}

We use a mobile robot dataset similar to the one reported in \cite{Tong2013}, collected in an indoor, planar environment featuring 17 tubular landmarks.
Both the robot and the tubes were equipped with reflective markers tracked by a ten-camera Vicon motion capture system.
This system provides sub-millimeter accuracy for the position of each reflective marker, which we use as ground truth for the robot's trajectory.
It further allows us to determine the exact location of the tubular landmarks, which we assume to be known during our localization experiment.

\begin{figure}[ht!]
	\centering
	\includegraphics[width=0.85\linewidth]{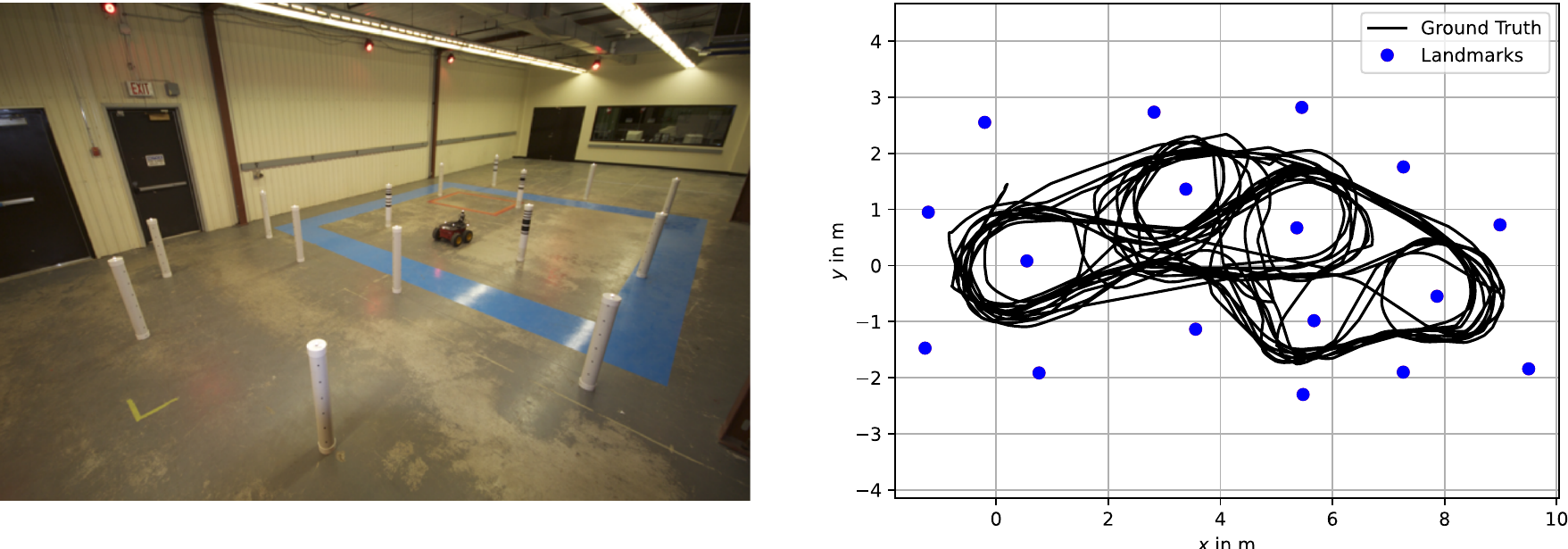}
	\caption{Left: Setup of the mobile robot localization experiment. Right: Ground-truth data of the mobile robot trajectory and landmark positions.}
	\label{fig:mobile_exp_setup}
\end{figure}

The mobile robot was driven around for 21 minutes.
The dataset includes laser rangefinder scans from a Hokuyo URG-04LX sensor, consisting of 681 range measurements spread over a 240$^\circ$ horizontal field of view centered straight ahead, logged at 10 Hz.
Additionally, wheel odometry readings were recorded at 10 Hz to provide estimates of the robot's body-centric velocity.
The experimental setup and ground-truth data of the recorded trajectory are shown in Fig.~\ref{fig:mobile_exp_setup}.
It can be seen that the robot’s path is quite twisted and looping.

\subsection{Localization Evaluation}

For evaluation, we consider the localization problem, where landmark locations are known. We are using the wheel odometry readings as piecewise-linear velocity inputs at a rate of 10 Hz to create a GP motion prior.
Landmark distance readings are included as discrete measurements, which are fused with the prior to create a posterior estimate.
We employ a MAP estimation scheme that computes the posterior distribution of the entire trajectory in a single batch.
We are particularly interested in how the proposed approach performs compared to the conventional WNOA GP prior baseline, in which odometry readings are considered as velocity measurements of the state, instead of as inputs.
We believe that this comparison is meaningful, as the WNOA prior is one of the most widely used and well-established methods for continuous-time state estimation \cite{Talbot2024}, which our proposed method directly extends.
While other methods, such as spline-based techniques, exist, they are fundamentally different from the GP-based methods considered in this work, with detailed comparisons reported in \cite{Johnson2024}.
For both our proposed method and the chosen baseline, we optimize the hyperparameters of the prior, namely $\mathbf{Q}_c$, on a subset of the available mobile robot driving data.
For the WNOA method, we additionally optimize the covariance of the odometry velocity measurements.
The covariance for the landmark distance measurements is given. 
The resulting hyperparameters used throughout the experiment are summarized in Table~\ref{tab:hyperparameters_mobile}.
Here, the prior covariances include one translational and one rotational degree of freedom, considering a non-holonomic mobile robot in a two-dimensional environment.
Since our derivations are formulated in $SE(3)$, the remaining degrees of freedom are locked during optimization.
Similarly, the velocity measurement covariance includes two values, as we measure the forward and yaw velocities of the mobile robot using wheel odometry readings.

\begin{table}[t]\renewcommand{\arraystretch}{1.5}
	\centering
	\caption{Hyperparameters used for mobile robot experiment}
	\label{tab:hyperparameters_mobile}
	\footnotesize
	\begin{tabular}{|c|c|c|c|}
		\hline \multicolumn{2}{|c|}{\textbf{Prior Covariance (Prior with Inputs)}}& \multicolumn{2}{|c|}{\textbf{Landmark Distance Measurement Covariance}} \\
		\hline \multicolumn{2}{|c|}{$\mathbf{Q}_{c} = \mbox{diag}\left(1.77\mathrm{e}{-5}~\mathrm{m}^2~3.50\mathrm{e}{-5}~\mathrm{rad}^2\right)$}& \multicolumn{2}{|c|}{$\mathbf{R}_\mathrm{dist} = 9.00\mathrm{e}{-4}~\mathrm{m}^2$} \\		
		\hline \hline \multicolumn{2}{|c|}{\textbf{Prior Covariance (Prior without Inputs)}}& \multicolumn{2}{|c|}{\textbf{Velocity Measurement Covariance}} \\
		\hline \multicolumn{2}{|c|}{$\mathbf{Q}_{c} = \mbox{diag}\left(2.11\mathrm{e}{-3}~\mathrm{m}^2~3.94\mathrm{e}{-2}~\mathrm{rad}^2\right)$}& \multicolumn{2}{|c|}{$\mathbf{R}_\mathrm{vel} = \mbox{diag}\left(5.45\mathrm{e}{-4}~\mathrm{m}^2/\mathrm{s}^2~1.01\mathrm{e}{-3}~\mathrm{rad}^2/\mathrm{s}^2\right)$} \\	\hline
	\end{tabular}
\end{table}

When estimating the robot's trajectory, we consider different frequencies in which landmark distance measurements are available to study increasingly sparse measurement availabilities.
For the initial evaluation, we include a discrete estimation state at each of the 10 Hz odometry updates into our batch optimization approach.
Afterwards, evaluation is repeated, while only including estimation states at times at which landmark distance measurements are available, meaning that less states are explicitly estimated as the landmark reading frequency decreases.
In these cases, odometry data is still considered at a rate of 10 Hz.
For the proposed method, the high-rate odometry information can be considered in a straightforward manner by constructing GP priors using piecewise-linear inputs between two estimation times.
For the WNOA GP prior baseline, high-rate velocity measurements are incorporated while relying on GP interpolation at runtime to associate them with the robot's state at the corresponding time (see \cite{Burnett2024a} for details).

In each case, the state is evaluated and compared to ground-truth at 10 Hz, meaning that we have to rely on GP interpolation to recover the posterior estimates at times between the discrete estimation times.
For the proposed approach, this interpolation considers the underlying motion prior with the incorporated inputs, allowing us to recover the local shape of the robot trajectory.
For the WNOA GP prior baseline, interpolation is solely based on the motion prior, in which the most likely robot state features a constant velocity.
In every case, we compute the root-mean-square errors (RMSE) for position and orientation between the ground-truth and estimated robot states.
We further record the computation time of the state estimation, using a C++ implementation of our method on a consumer laptop featuring a 2.4 GHz Intel i5 processor with 16 GB of RAM.

\begin{table}[t]\renewcommand{\arraystretch}{1.75}
	\setlength{\tabcolsep}{5pt}
	\centering
	\footnotesize
	\caption{Mobile robot trajectory estimation results considering odometry readings as measurements vs.\ as inputs}
	\label{tab:mobile_results}
	% First part of the table
	\begin{tabular}{|r||cc|cc|c||cc|cc|c|}
		\hline
		& \multicolumn{5}{c||}{\textbf{Prior Without Inputs}}
		& \multicolumn{5}{c|}{\textbf{Prior With Inputs (Ours)}} \\
		\hline
		\multirow{2}{*}{\makecell{$\boldsymbol{\Delta}\boldsymbol{t_l}$ \\ \textbf{in s}}}
		& \multicolumn{2}{c|}{\textbf{Pos.\ Error (cm)}}
		& \multicolumn{2}{c|}{\textbf{Rot.\ Error (°)}}
		& \multirow{2}{*}{\makecell{\textbf{Time} \\ \textbf{in s}}}
		& \multicolumn{2}{c|}{\textbf{Pos.\ Error (cm)}}
		& \multicolumn{2}{c|}{\textbf{Rot.\ Error (°)}}
		& \multirow{2}{*}{\makecell{\textbf{Time} \\ \textbf{in s}}} \\
		\cline{2-5} \cline{7-10}
		& \textbf{RMSE} & \textbf{Max}
		& \textbf{RMSE} & \textbf{Max}
		&
		& \textbf{RMSE} & \textbf{Max}
		& \textbf{RMSE} & \textbf{Max}
		& \\
		\hline \hline
		{0.5}
		& \textbf{2.44} & \textbf{16.37} & \textbf{4.87}  & \textbf{15.15} & 12.9
		& \textbf{2.59} & \textbf{15.71} & \textbf{4.74} & \textbf{15.14} & 21.9 \\
		{1}
		& \textbf{2.49} & \textbf{17.29} & \textbf{4.90} & \textbf{15.12} & 10.6
		& \textbf{2.61} & \textbf{16.10} & \textbf{4.73} & \textbf{15.15} & 21.1 \\
		{2}
		& \textbf{2.70}  & \textbf{17.98} & \textbf{4.99} & \textbf{15.34} & 10.8
		& \textbf{2.76} & \textbf{16.72} & \textbf{4.77} & \textbf{15.17} & 18.3 \\
		{3}
		& \textbf{2.93} & \textbf{17.44} & \textbf{5.00} & \textbf{15.58} & 10.6
		& \textbf{2.99} & \textbf{15.99} & \textbf{4.74} & \textbf{15.70} & 20.8 \\
		{4}
		& \textbf{3.20}  & \textbf{18.36} & \textbf{5.03} & \textbf{15.96} & 9.1
		& \textbf{3.10}  & \textbf{17.10} & \textbf{4.76}  & \textbf{15.68} & 19.6 \\
		{5}
		& \textbf{3.48} & \textbf{19.54} & \textbf{5.04}  & \textbf{16.88} & 9.1
		& \textbf{3.09} & \textbf{17.82} & \textbf{4.75} & \textbf{17.10} & 21.1 \\
		{6}
		& \textbf{3.58} & \textbf{18.45} & \textbf{5.08} & \textbf{16.06} & \textbf{10.8}
		& \textbf{3.47} & \textbf{17.19} & \textbf{4.80} & \textbf{16.24} & 23.9 \\
		{7}
		& \textbf{4.13} & \textbf{24.87} & \textbf{5.20} & \textbf{16.59} & \textbf{10.4}
		& \textbf{3.71} & \textbf{25.00} & \textbf{4.86} & \textbf{17.02} & 20.8 \\
		\hline\hline
		& \multicolumn{5}{c||}{\makecell{\textbf{Prior Without Inputs} \\ \textbf{(Exclude states w/o dist.\ meas.)}}}
		& \multicolumn{5}{c|}{\makecell{\textbf{Prior With Inputs (Ours)} \\ \textbf{(Exclude states w/o dist.\ meas.)}}} \\
		\hline
		\multirow{2}{*}{\makecell{$\boldsymbol{\Delta}\boldsymbol{t_l}$ \\ \textbf{in s}}}
		& \multicolumn{2}{c|}{\textbf{Pos.\ Error (cm)}}
		& \multicolumn{2}{c|}{\textbf{Rot.\ Error (°)}}
		& \multirow{2}{*}{\makecell{\textbf{Time} \\ \textbf{in s}}}
		& \multicolumn{2}{c|}{\textbf{Pos.\ Error (cm)}}
		& \multicolumn{2}{c|}{\textbf{Rot.\ Error (°)}}
		& \multirow{2}{*}{\makecell{\textbf{Time} \\ \textbf{in s}}} \\
		\cline{2-5} \cline{7-10}
		& \textbf{RMSE} & \textbf{Max}
		& \textbf{RMSE} & \textbf{Max}
		&
		& \textbf{RMSE} & \textbf{Max}
		& \textbf{RMSE} & \textbf{Max}
		& \\
		\hline \hline
		{0.5}
		& \textbf{2.48} & \textbf{15.96} & \textbf{4.85} & \textbf{14.63} & \textbf{3.2}
		& \textbf{2.59} & \textbf{15.77} & \textbf{4.74} & \textbf{15.14} & \textbf{3.1} \\
		{1}
		& \textbf{2.84} & \textbf{17.09} & \textbf{5.00} & \textbf{15.57} & \textbf{2.3}
		& \textbf{2.63} & \textbf{16.35} & \textbf{4.73} & \textbf{15.18} & \textbf{2.5} \\
		{2}
		& 4.24         & 13.80          & 5.71          & 19.16          & 2.3
		& \textbf{2.85} & \textbf{17.33} & \textbf{4.77} & \textbf{15.32} & \textbf{2.8} \\
		{3}
		& 6.95         & 27.17          & 6.50          & 25.88          & 2
		& \textbf{3.17} & \textbf{16.45} & \textbf{4.76}  & \textbf{15.88} & \textbf{3.7} \\
		{4}
		& 9.56         & 32.04          & 8.07          & 33.97          & 1.8
		& \textbf{3.53} & \textbf{17.86} & \textbf{4.79} & \textbf{16.12} & \textbf{4.5} \\
		{5}
		& 13.08         & 47.20          & 9.35          & 41.73          & 1.8
		& \textbf{3.68} & \textbf{17.83} & \textbf{4.80} & \textbf{17.48} & \textbf{6} \\
		{6}
		& 17.09         & 60.60          & 10.93         & 45.66          & 1.7
		& 4.64         & 16.93          & 4.89          & 16.49          & 7.8 \\
		{7}
		& 22.83         & 71.80          & 13.10         & 52.46          & 2.2
		& 5.85         & 23.55          & 4.90          & 16.58          & 10.1 \\
		\hline
	\end{tabular}
\end{table}

\subsection{Results}

The results are summarized in Tab.~\ref{tab:mobile_results}, including the estimation errors and computation times for increasing times between landmark distance measurements $\Delta t_l$.
When including a discrete state at every odometry update, both methods achieve similar results.
The resulting accuracies differ only by a few mm and less than half of a degree, while the WNOA GP prior baseline achieves shorter computation times.
In each case shown, the 21-minute trajectory can be estimated in just a few seconds.
When including discrete states only when landmark distance measurements occur, the proposed prior formulation using inputs significantly outperforms the baseline.
While the traditional WNOA GP prior is able to achieve competitive accuracies for short time distances $\Delta t_l$, it fails to accurately represent the local shape of the trajectory when landmark measurements occur less frequently ($\Delta t_l > 1$ s).
The proposed GP prior with inputs is able to maintain the estimation accuracy for longer times.
Here the error only starts to significantly increase for $\Delta t_l > 5$ s.
Both methods show significantly reduced computation times when excluding states without landmark distance measurements in the batch estimation.
However, the computation time increases when using a prior with inputs as $\Delta t_l$ increases.
In these cases, more states have to be recovered using the GP interpolation schemes, which requires additional effort for the proposed method to solve the Magnus expansion and integral terms.

\begin{figure}[ht!]
	\centering
	\includegraphics[width=0.8\linewidth]{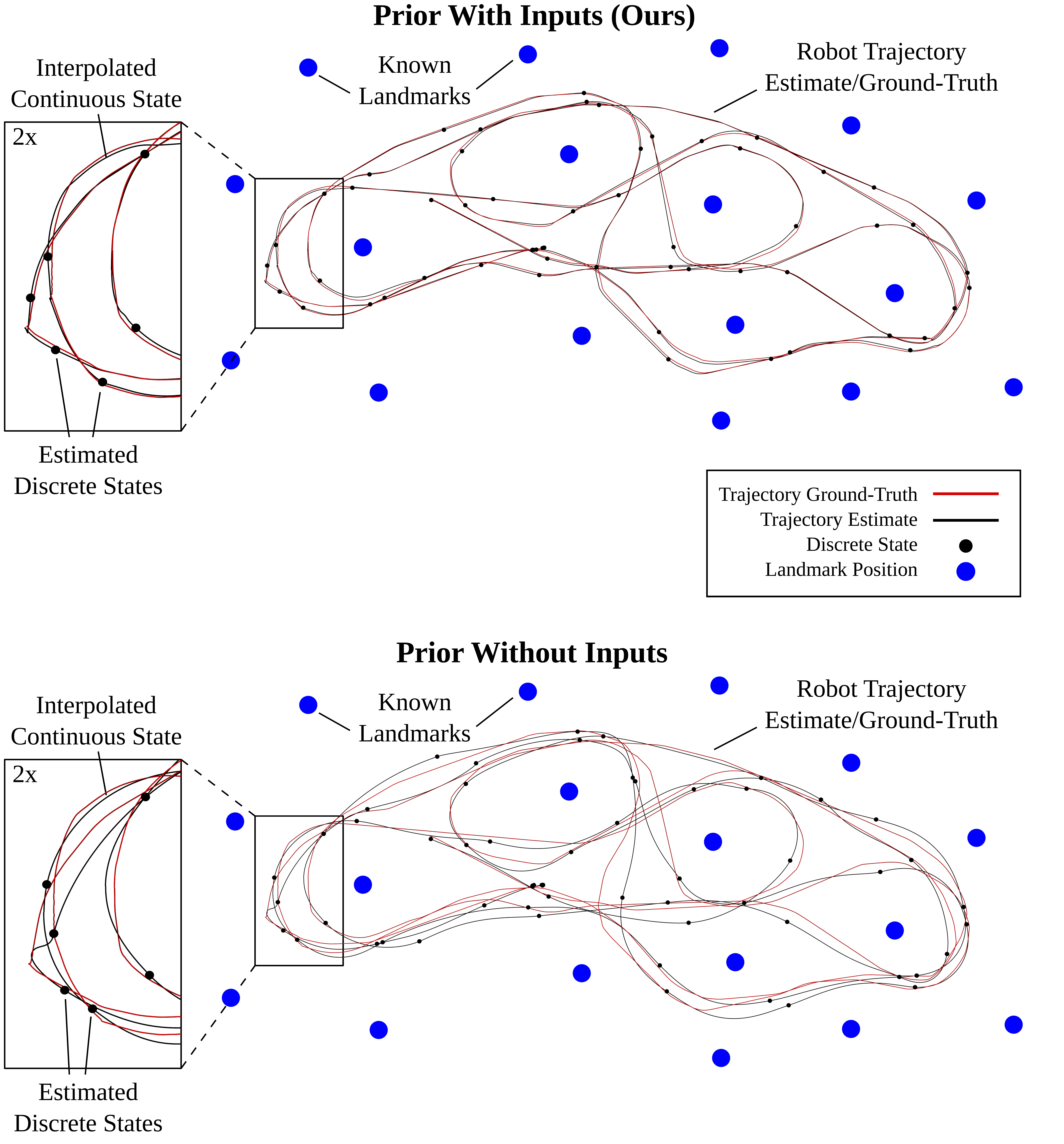}
	
	\caption{Example state estimation results using the first 25\% of the mobile robot dataset. Trajectory estimates are depicted in black with ground-truth in red. Landmarks are shown in blue. The discrete robot state is estimated every 5 seconds, coinciding with landmark distance measurements, while the continuous state relies on interpolation. On the top, the results for the newly proposed state estimation method are shown, using the odometry readings as velocity inputs. On the bottom, the conventional WNOA GP prior is used, considering the odometry as velocity measurements instead}
	\label{fig:mobile_exp}
\end{figure}

We conclude that using the proposed GP prior with inputs presents an interesting use case, as it is able to maintain the estimation accuracy, while significantly reducing the computational effort.
This is thanks to its ability to maintain the local shape of the odometry over longer time intervals without needing to include discrete states in the batch estimation.
We can exploit this advantage, as long as $\Delta t_l$ does not grow too large, in which case the accuracy and computation times are not competitive with the WNOA GP prior baseline, even without excluding discrete states from the batch estimation.
In our case, it is beneficial to use the newly proposed method for $\Delta t_l \leq 5$ s, excluding states without landmark distance measurements.
For $\Delta t_l > 5$ s the original WNOA method should be applied, including all discrete states and treating odometry velocities as measurements.

To further illustrate the differences of the two approaches, Fig.~\ref{fig:mobile_exp} shows state estimation results for the first 25\% of the mobile robot dataset, while considering landmark distance measurements every $\Delta t_l = 5$ s and including discrete estimation states only at these measurement times.
The proposed method is able to accurately represent the local shape of the trajectory between the discrete states.
In contrast, the baseline, which relies on the constant-velocity prior, cannot accurately capture the local shape of the trajectory and instead produces a highly smoothed result.
This leads to considerable inaccuracies in the resulting state estimation.
This is particularly evident in the zoomed-in area of the trajectory, which includes a sharp turn. Here, the robot stopped and pivoted on the spot between two discrete estimation times, resulting in a non-smooth trajectory. This motion is captured by the prior that considers velocities as inputs, but cannot be represented by the WNOA prior, whose interpolation overly smooths the trajectory, deviating from the ground-truth.

\section{Continuum Robot Shape Estimation}

As discussed in detail in \cite{Lilge2022}, the GP batch state estimation approach from \cite{Anderson2015} for mobile robot trajectories can also be applied to continuously estimate quasi-static continuum robot shapes. Consequently, the proposed GP prior that incorporates inputs can be extended to this domain as well.
Analogous to \eqref{eq:motion_ode}, we can describe the continuum robot shape along its arclength $s$ with the following stochastic differential equation:
\begin{subequations}
	\begin{align} \label{eq:shape_ode}
		\frac{d}{ds}{\mbf{T}}(s) &= (\underbrace{\mbs{\varepsilon}_\mathrm{bias}(s) + \mbs{\varepsilon}_\mathrm{in}(s)}_{\mbs{\varepsilon}(s)})^\wdg\mbf{T}(s), \\
		\frac{d}{ds}{\mbs{\varepsilon}}_\mathrm{bias}(s) &= \mbs{\mathcal{K}}^{-1}\mbs{f}_\mathrm{in}(s) + \mbf{w}(s), \\ \mbf{w}(s) &\sim \mathcal{GP}(\mbf{0},\mbf{Q}_c(s-s')).
	\end{align}
\end{subequations}
These equations describe the evolution of the robot's pose $\mbf{T}(s) \in SE(3)$ and strain $\mbs{\varepsilon}(s) \in \mathds{R}^6$ expressed in the body frame along its arclength.
Together, they define the quasi-static shape of the continuum robot.
Similar to the mobile robot equations, we consider two potential inputs: an input strain $\mbs{\varepsilon}_\mathrm{in}(s) \in \mathds{R}^6$ and distributed forces and moments $\mbs{f}_\mathrm{in}(s) \in \mathds{R}^6$, which can act as an input to the system.
These distributed loads are mapped to the strain derivative through the stiffness matrix $\mbs{\mathcal{K}}$ of the continuum robot's backbone.
Lastly, $\mbf{w}(s)$ is again a white-noise GP with positive-definite power-spectral density matrix $\mbf{Q}_c$.
Using this expression and the previous derivations, we can derive an analogous GP prior with inputs for the continuous estimation of the quasi-static shape of continuum robots.

Throughout this section, we validate this GP prior formulation with additional continuum robot experiments.

\begin{figure}[ht]
	\centering
	\includegraphics[width=0.25\linewidth]{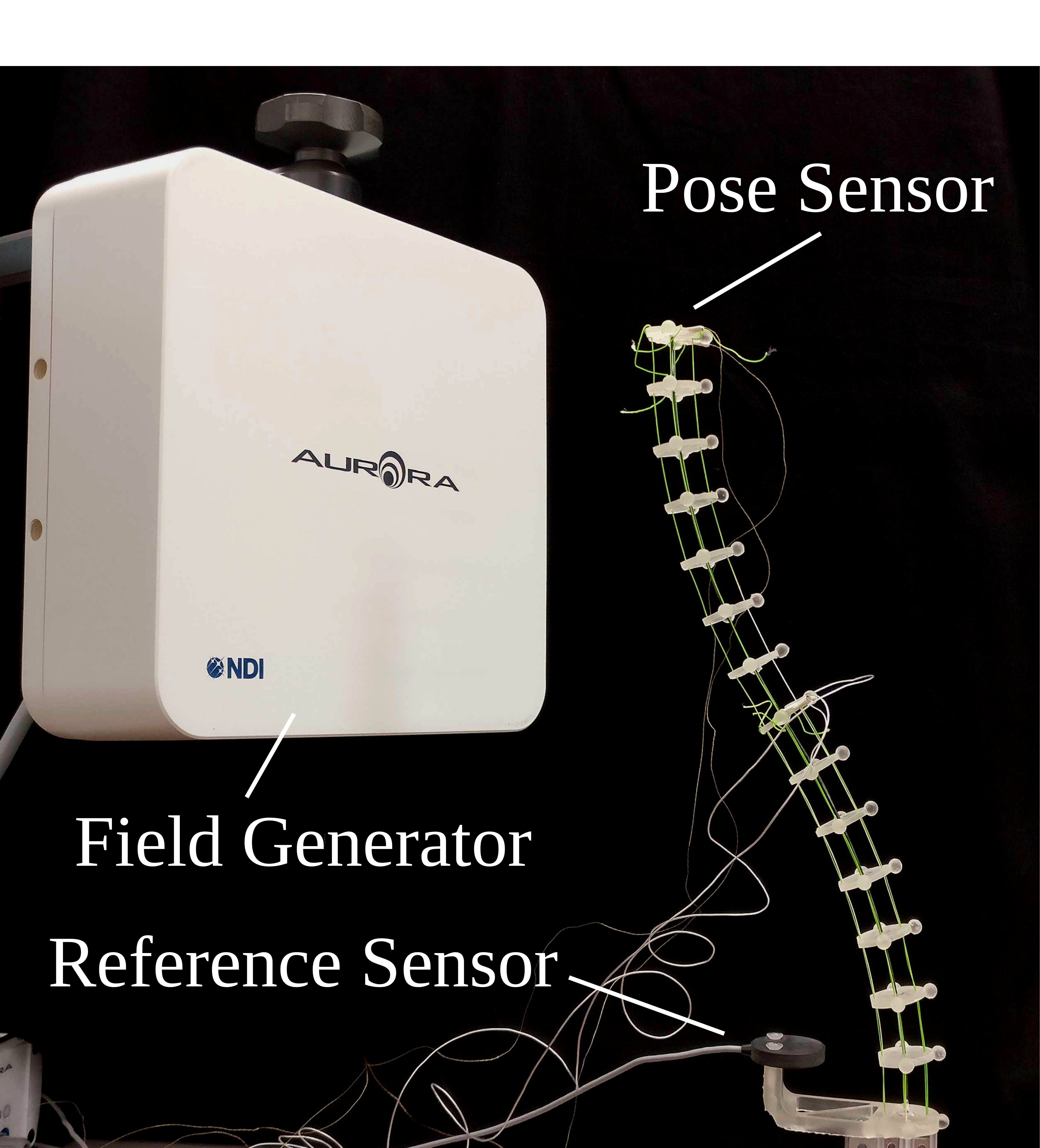}	
	\caption{Tendon-driven continuum robot prototype and experimental setup}
	\label{fig:continuum_setup}
\end{figure}

\subsection{Experimental Setup}

For the experiments, we are using the tendon-driven continuum robot (TDCR) prototype and sensor setup presented in \cite{Lilge2022} (see Fig.~\ref{fig:continuum_setup}).
This prototype features two controllable segments, with tendons routed parallel to the backbone through spacer disks and attached at the end of each segment.
The bending of both segments can be individually actuated by pulling and releasing the tendons.
For sensing, we consider a single six-degree-of-freedom pose measurement at the robot's tip using an electromagnetic tracking sensor (Aurora v3, Northern Digital Inc., Canada).
Ground-truth is obtained using a coordinate measurement arm with an attached laser probe (FARO Edge with FARO Laser Line Probe HD, FARO Technologies Inc., USA).
This measurement arm allows us to capture a detailed laser scan of the robot's shape in the form of a point cloud, from which the discrete pose of each spacer disk can be reconstructed.
Six simple bending configurations of the continuum robot were exclusively recorded to calibrate the sensor and ground-truth frames.
Using the same procedure outlined in \cite{Lilge2022}, the remaining calibration error at the tip of the robot results in $3.3 \pm 1.1$~mm.

\begin{figure}[t]
	\centering
	\includegraphics[width=1\linewidth]{figs/continuum_exp_new.pdf}	
	\caption{Example continuum robot state estimation results using prior formulations without and with known actuation inputs. Results are presented for both the full pose measurement (top) and the position-only measurement (bottom) of the continuum robot's tip}
	\label{fig:continuum_exp}
\end{figure}

\subsection{Evaluation}

For evaluation, we consider nine different TDCR configurations, with three sets of applied tendon tensions. Each set of tensions is shared across three configurations.
In each configuration, an additional unknown force is applied to the robot's tip to induce further deflection.

For state estimation, we now wish to compare the conventional prior formulation from \cite{Lilge2022} with the newly proposed one considering inputs. 
As discussed above, there are two possibilities to incorporate known inputs.
One can either consider known input strains $\mbs{\varepsilon}_\mathrm{in}(s)$ or known distributed actuation loads $\mbs{f}_\mathrm{in}(s)$.
Either of these possible inputs can, for instance, be provided by another open-loop continuum robot model.

We use the second approach and convert the known applied tendon tensions to moments acting on the robot backbone, considering the simplified point-moment model discussed in \cite{Rucker2011}.
Using this model, we compute the discrete moments applied to the tendon termination points.
We then approximate these discrete moments with piecewise-linear inputs to obtain an expression for $\mbs{f}_\mathrm{in}(s)$, which we map to the strain derivative using the known stiffness $\mbc{\mathcal{K}}$ of the robot's Nitinol backbone, based on properties reported by the manufacturer. 
While this simplified model exhibits inaccuracies in out-of-plane bending scenarios \cite{Rucker2011}, we deem it sufficiently accurate to be fused with additional measurements.
Alternatively, more complex actuation load distributions could be incorporated, as long as they can be expressed in the robot's body frame independently from the state variables.

In the mobile robot experiments, velocity inputs were included as measurements in the traditional WNOA method for a fairer comparison.
However, we do not include the inputs as measurements for the baseline method in the continuum robot evaluations. This is because our inputs are defined on the strain derivative, which is not part of the system state, making it difficult to directly include such measurements in a straightforward manner.

Both the traditional prior from \cite{Lilge2022} as well as the newly proposed prior considering inputs are fused with a single pose measurement at the tip of the robot.
As an additional degenerate case, we further consider the scenario in which only the robot tip's position can be measured.
For every case, we compute the RMSE and maximum errors for the robot's position and orientation along its shape, considering the discrete ground-truth poses at each disk.
Hyperparameters are tuned empirically on representative continuum robot configurations.
For both methods, the prior covariance is set to $\mathbf{Q}_{c}~=~\mbox{diag}\left(1\mathrm{e}{-2}~\mathrm{m}^2~1\mathrm{e}{-2}~\mathrm{m}^21\mathrm{e}{-2}~\mathrm{m}^2~1\mathrm{e}{3}~\mathrm{rad}^2~1\mathrm{e}{3}~\mathrm{rad}^2~1\mathrm{e}{3}~\mathrm{rad}^2\right)$, while the pose measurement covariance is $\mathbf{R}_\mathrm{pose}~=~\mbox{diag}\left( 4\mathrm{e}{-7}~\mathrm{m}^2~4\mathrm{e}{-7}~\mathrm{m}^2~4\mathrm{e}{-7}~\mathrm{m}^2~2.5\mathrm{e}{-4}~\mathrm{rad}^2~2.5\mathrm{e}{-4}~\mathrm{rad}^2~2.5\mathrm{e}{-4}~\mathrm{rad}^2\right)$.

\subsection{Results}

The results for each of the nine configurations, as well as their average, are reported in Tab.~\ref{tab:continuum_robot_experiments_results}.
In each case, the state estimation problem could be solved in 5-10 ms.
Fig.~\ref{fig:continuum_exp} additionally showcases the state estimation results for the first configuration, both when considering a full pose measurement and a position-only measurement of the continuum robot's tip.
It can be seen that due to the ability to retain the local shape of the prior, our proposed method outperforms the conventional GP formulation in every case, with two exceptions.
In these cases, relatively high external forces were applied to the continuum robot, which resulted in shapes that significantly differ from the utilized prior, making it less helpful.
Nevertheless, the results in these cases remain largely comparable to those obtained by the baseline method.

\begin{table}[t]\renewcommand{\arraystretch}{1.4}
	\centering
	\footnotesize
	\setlength{\tabcolsep}{3pt}
	\caption{Continuum robot shape estimation results using priors with and without known actuation inputs}
	\label{tab:continuum_robot_experiments_results}
	\begin{tabular}{|c||c|c|c|c||c|c|c|c||c|c|c|c||c|c|c|c|}
		\hline 
		& \multicolumn{8}{c||}{\textbf{Pose Measurement}} & \multicolumn{8}{c|}{\textbf{Position Measurement}} \\
		\hline & \multicolumn{4}{c||}{\textbf{Prior Without Inputs}} & \multicolumn{4}{c||}{\textbf{Prior With Inputs (Ours)}} & \multicolumn{4}{c||}{\textbf{Prior Without Inputs}} & \multicolumn{4}{c|}{\textbf{Prior With Inputs (Ours)}} \\
		
		\hline & \multicolumn{2}{c|}{Pos. in mm} & \multicolumn{2}{c||}{Rot. in $^\circ$} & \multicolumn{2}{c|}{Pos. in mm} & \multicolumn{2}{c||}{Rot. in $^\circ$} & \multicolumn{2}{c|}{Pos. in mm} & \multicolumn{2}{c||}{Rot. in $^\circ$} & \multicolumn{2}{c|}{Pos. in mm} & \multicolumn{2}{c|}{Rot. in $^\circ$} \\
		\hline \textbf{Config} & RMSE & Max & RMSE & Max & RMSE & Max & RMSE  & Max & RMSE & Max & RMSE & Max & RMSE & Max & RMSE  & Max \\
		\hline \hline 1 & 10.5 & 16.3 & 6.5 & 10.9 & \textbf{2.9} & \textbf{4.7} & \textbf{3.2}  & \textbf{5.8} & 14.1 & 24.1 & 14.5 & 32.0 & \textbf{1.7} & \textbf{2.7} & \textbf{4.4}  & \textbf{6.6} \\
		2 & 7.6 & 13.2 & 7.7 & 13.1 & \textbf{3.3} & \textbf{5.7} & \textbf{6.8}  & \textbf{11.7} & 18.6 & 30.8 & 22.1 & 32.8 & \textbf{3.8} & \textbf{5.7} & \textbf{10.0}  & \textbf{14.1} \\
		3 & 12.0 & 17.7 & 5.8 & 10.2 & \textbf{4.7} & \textbf{7.3} & \textbf{3.7}  & \textbf{6.3} & 11.3 & 20.0 & 13.4 & 31.0 & \textbf{5.2} & \textbf{7.9} & \textbf{6.5}  & \textbf{10.5} \\
		4 & 9.2 & 12.7 & 6.4 & 12.0 & \textbf{5.2} & \textbf{10.2} & \textbf{6.0}  & \textbf{11.1} & \textbf{5.2} & \textbf{9.3} & 11.6 & 20.6 & 9.2 & 13.3 & \textbf{10.5}  & \textbf{15.4} \\
		5 & \textbf{3.6} & \textbf{5.7} & 6.8 & 12.4 & 6.5 & 8.9 & \textbf{6.1}  & \textbf{8.5} & 18.8 & 29.6 & 37.2 & 48.1 & \textbf{5.4} & \textbf{7.1} & \textbf{12.5}  & \textbf{16.0} \\
		6 & 8.3 & 12.7 & 7.0 & 11.5 & \textbf{6.3} & \textbf{8.4} & \textbf{3.3}  & \textbf{4.9} & 7.8 & 12.5 & 14.2 & 20.6 & \textbf{6.7} & \textbf{9.5} & \textbf{5.2}  & \textbf{7.8} \\
		7 & 9.0 & 13.5 & 7.0 & 11.7 & \textbf{3.0} & \textbf{5.0} & \textbf{4.2}  & \textbf{6.7} & 16.5 & 27.4 & 16.1 & 34.5 & \textbf{6.9} & \textbf{11.3} & \textbf{14.7}  & \textbf{21.9} \\
		8 & 11.4 & 17.3 & 11.4 & 19.8 & \textbf{2.0} & \textbf{3.2} & \textbf{6.2}  & \textbf{9.7} & 13.8 & 23.5 & 21.9 & 38.2 & \textbf{8.4} & \textbf{13.7} & \textbf{21.4}  & \textbf{27.4} \\
		9 & 10.7 & 16.0 & 10.0 & 15.7 & \textbf{4.4} & \textbf{6.8} & \textbf{8.5}  & \textbf{13.6} & 19.7 & 32.9 & 42.6 & 57.2 & \textbf{6.3} & \textbf{9.5} & \textbf{11.0}  & \textbf{17.9} \\
		\hline\hline Avg. & 9.1 & 13.9 & 7.6 & 13.0 & \textbf{4.2} & \textbf{6.7} & \textbf{5.3}  & \textbf{8.7} & 14.0 & 23.3 & 21.5 & 35 & \textbf{6.0} & \textbf{9.0} & \textbf{10.7}  & \textbf{15.3} \\
		\hline
	\end{tabular}
\end{table}

In the remaining configurations, the accuracy of the proposed method is significantly higher than that of the baseline method, especially in cases where only the position of the robot's tip is measured. 
In these instances, the baseline method's shape estimates largely deviate from the ground-truth, with worst-case maximum errors exceeding 3 cm and 55$^\circ$. This further underscores the proposed method's effectiveness in situations with sparse sensing, as the additional inputs contribute to a more informed prior, resulting in more a accurate state estimation.
This is also apparent in the example shown in Fig.~\ref{fig:continuum_exp}.

\section{Conclusion}

In this work, we propose an extension to the WNOA GP prior used in continuous-time batch state estimation.
Our new formulation allows us to incorporate both known velocity and acceleration inputs, as long as they can be expressed in the robot's body frame independently of the robot state.
We specifically present the case of piecewise-linear inputs, for which closed-form approximations can be obtained to represent motions in $SE(3)$.
Experiments showcase the use cases of the method for both mobile robot trajectory and continuum robot shape estimation.
It is shown that considering inputs in the GP prior is particularly helpful to maintain the local shape of the trajectory between discrete state observations, specifically when the motion differs from the constant-velocity assumption of the traditional WNOA GP prior.

The main advantage of the proposed approach is the introduction of more informed priors, potentially reducing the number of measurements and estimation nodes required to achieve accurate state estimates.
This reduction can significantly decrease the computation time for state estimation problems, as shown during mobile robot experiments.
Furthermore, as shown during the continuum robot experiments, the approach can enable more accurate estimations in scenarios with limited sensing, as the inclusion of control inputs adds valuable information to the prior, compensating for the lack of measurements.

Nevertheless, the proposed method has some limitations.
Incorporating control inputs increases the complexity of computing the prior and posterior mean and covariances.
As a result, both the MAP optimization process and state querying require more computation time compared to the baseline method.
Favourable computation times can only be achieved when using less frequent estimation nodes, where the proposed approach can still maintain estimation accuracy.
Another limitation arises when the control inputs exhibit discontinuities, which persist in the posterior distribution and reduce its smoothness.
While this characteristic may be advantageous in certain scenarios, it can negatively impact accuracy when noisy measurements are directly used as inputs.
Furthermore, for mobile robots, the presented experiments did not demonstrate a significant improvement in accuracy when incorporating velocities as inputs rather than as measurements, particularly with frequent discrete estimation nodes.
This may be due to the fact that the traditional WNOA prior already models the robot's motion quite effectively, assuming constant velocity between discrete time steps.

Future work will focus on evaluating the proposed approach in scenarios involving more aggressive motions, such as drone racing, where the traditional WNOA GP prior is less effective.
This will help identify situations in which incorporating control inputs into GP priors is particularly advantageous.
Additionally, we aim to investigate whether similar benefits can be achieved by incorporating inputs into the WNOJ GP motion prior as an alternative.

\section*{Acknowledgements}

The authors thank the Continuum Robotics Laboratory at the University of Toronto for providing the prototype and facilities for conducting the continuum robot experiment.

\section*{Author Contribution}
SL and TDB both conceptualized the research question, method and experiments. SL wrote the majority of the paper, derived and implemented the proposed method and carried out the qualitative and quantitative evaluations. TDB is the senior author on this paper. He regularly participated in discussions on the proposed research, method and results, and edited the paper.

\section*{Financial Support}
This work was supported by the Natural Sciences and Engineering Research Council of Canada (NSERC).

\section*{Competing Interests Declaration}

The authors declare none.

\bibliographystyle{roblike}
\bibliography{refs}

\end{document}

%% file: notation_header.tex
\newcommand{\bbm}{\begin{bmatrix}}
\newcommand{\ebm}{\end{bmatrix}}
\newcommand{\mbf}{\mathbf}
\newcommand{\mbs}[1]{{\boldsymbol{#1}}}

\newcommand{\beq}{\begin{equation}}
\newcommand{\eeq}{\end{equation}}
\newcommand{\bdis}{\begin{displaymath}}
\newcommand{\edis}{\end{displaymath}}
\newcommand{\beqn}[1]{\begin{subequations}\label{eq:#1}\begin{eqnarray}}
\newcommand{\eeqn}{\end{eqnarray}\end{subequations}}

\newcommand{\pri}[1]{\check{#1}}
\newcommand{\wdg}{\wedge}

\newcommand{\Wdg}{\curlywedge}

\newcommand{\mbc}[1]{{\mathcal{#1}}}